%% file: samplepaper.tex
\documentclass[runningheads]{llncs}
\usepackage[T1]{fontenc}
\usepackage{graphicx}
\usepackage{amsmath}
\usepackage{subcaption}
\usepackage{booktabs}
\usepackage{multirow} 
\usepackage{algorithm}
\usepackage{algpseudocode}

\usepackage{url}
\usepackage{hyperref}
\usepackage{orcidlink}

\usepackage{amssymb}

\newif\ifanonymous
\anonymousfalse   

\begin{document}
\setlength{\textfloatsep}{10pt}
\setlength{\floatsep}{10pt}
\title{PerfMamba: Performance Analysis and Pruning of
Selective State Space Models}
\titlerunning{PerfMamba}
%
\ifanonymous
  \author{Anonymous Authors}
  \institute{Submission id: 26}

\else

   \author{
  Abdullah Al Asif\orcidlink{0000-0002-0389-931X}\inst{1}\thanks{Corresponding authors.} \and
  Mobina Kashaniyan\orcidlink{0009-0000-4332-8442}\inst{1} \and
  Sixing Yu\orcidlink{0000-0002-2415-566X}\inst{1} \and
  Juan Pablo Muñoz\inst{2} \and
  Ali Jannesari\orcidlink{0000-0001-8672-5317}\inst{1}\protect\footnotemark[1]
}

  \authorrunning{A.\ A.\ Asif et al.}

  \institute{
  Iowa State University, Ames, Iowa, USA \\
  \email{\{aaasif, mobina, yusx, jannesar\}@iastate.edu}
  \and
  Intel Labs, USA \\
  \email{jpablomch@gmail.com, pablo.munoz@intel.com}
}
\fi
\maketitle              
\begin{abstract}
Recent advances in sequence modeling have introduced selective SSMs as promising alternatives to Transformer architectures, offering theoretical computational efficiency and sequence processing advantages. A comprehensive understanding of selective SSMs in runtime behavior, resource utilization patterns, and scaling characteristics still remains unexplored, thus obstructing their optimal deployment and further architectural improvements. This paper presents a thorough empirical study of Mamba-1 and Mamba-2, systematically profiled for performance to assess the design principles that contribute to their efficiency in state-space modeling. A detailed analysis of computation patterns, memory access, I/O characteristics, and scaling properties was performed for sequence lengths ranging from 64 to 16384 tokens. Our findings show that the SSM component, a central part of the selective SSM architecture, demands a significant portion of computational resources compared to other components in the Mamba block.   Based on these insights, we propose a pruning technique that selectively removes low-activity states within the SSM component, achieving measurable throughput and memory gains while maintaining accuracy within a moderate pruning regime. This approach results in performance improvements across varying sequence lengths, achieving a 1.14x speedup and reducing memory usage by 11.50\%. These results offer valuable guidance for designing more efficient SSM architectures that can be applied to a wide range of real-world applications.

\keywords{Selective State Space Models (SSMs) \and Model Profiling \and Performance Optimization.}
\end{abstract}

\input{content/02_introduction}
\input{content/003_background}
\input{content/03_related_work}
\input{content/04_method}
\input{content/05_SSM}
\input{content/06_eval}

\input{content/07_future_work}

\input{content/08_ack}

\newpage
\bibliographystyle{splncs04}
\bibliography{reference}

\end{document}

%% file: content/02_introduction.tex
\section{Introduction}

Large language models (LLMs) have revolutionized natural language processing, yet their computational efficiency remains a critical challenge. While Transformer-based architectures dominate the field\cite{vaswani2023attentionneed}, their quadratic attention mechanism has spurred research into alternative architectures. Selective State Space Models (SSMs) have emerged as a promising direction, offering theoretical advantages through their linear scaling properties and efficient sequence modeling capabilities\cite{gu2024mambalineartimesequencemodeling}.

SSMs represent a fundamental shift in sequence modeling, replacing attention mechanisms with state space dynamics that capture long-range dependencies and local patterns. Recent implementations, including Mamba-1, Mamba-2\cite{dao2024transformersssmsgeneralizedmodels}, and other variants, have demonstrated 5x higher throughput and 1.7x lower memory usage compared to similarly sized Transformer models during inference\cite{gu2024mambalineartimesequencemodeling}.
This property makes them attractive for applications in natural language processing, speech, and time series, where long-range dependencies are common. Despite their growing adoption, however, the computational behavior of SSMs remains underexplored. In particular, it is unclear which internal components dominate runtime costs and how these costs scale with sequence length, leaving open questions about where to direct optimization efforts. 

In this work, we present \textbf{PerfMamba}, a study of the computational characteristics of Mamba-1 and Mamba-2. We profile their components across a range of sequence lengths (64--16k) and identify the SSM update as the significant cost driver, consistently accounting for more than half of the total computation and memory. This finding highlights a specific target to improve the efficiency of SSMs without altering their expressive power. Motivated by this observation, we propose \emph{$\Delta$-guided structured state pruning}, a novel method that exploits a signal unique to SSMs. Each Mamba layer computes an input-dependent continuous-time gate, $\Delta$, which modulates how strongly state channels retain or overwrite information. We show that averaging $\Delta$ values across data provides a simple but effective activity measure for each state channel. By removing channels with consistently low activity and introducing a lightweight bridging linear to preserve compatibility, we reduce the dimensionality of the SSM state space in a structured way. Unlike weight sparsification or unstructured pruning used in other architectures, our method directly targets the state dimension of SSMs, making it both principled and architecture-specific.

We evaluated this technique on Mamba-2-130M with sequence lengths up to 16k. Our experiments show that $\Delta$-guided pruning improves throughput and reduces memory footprint at long sequences. Accuracy remains stable in a moderate pruning regime ($\leq 30\%$ of states pruned), while more aggressive pruning exposes clear trade-offs across tasks. Beyond efficiency, this analysis also reveals how state activity is distributed across layers and datasets, providing additional interpretability into SSM dynamics.\\

\noindent This paper makes three key contributions:
\begin{enumerate}
    \item We conduct the first component-level profiling study of Mamba-1 and Mamba-2, identifying detailed resource consumption patterns for the core components of Mamba blocks. This breakdown provides actionable insights to guide optimization priorities and highlights components requiring further enhancements to improve both temporal and spatial efficiency.
    \item We present empirically derived best practices for deploying State Space Models (SSMs), showing how effective hardware--software co-optimization can preserve accuracy while delivering substantial efficiency gains across diverse sequence lengths.
    \item Building on these insights, we demonstrate that the proposed optimizations improve inference throughput by up to $1.14\times$ compared to baseline implementations, while also achieving memory savings of up to $11.5\%$ in the state update phase for long sequences.
    
\end{enumerate}

The remainder of this paper is organized as follows: Section \ref{sec:background} and \ref{sec:related_work} provide the necessary background on SSMs and related work in sequence model profiling. Section \ref{sec:profiling} describes our profiling methodology and presents detailed findings and analysis. Section \ref{sec:pruning} discusses the implications of our profiling results and presents an efficient pruning method for the state of the SSM, along with its implementation. Section \ref{sec:results} presents the results and evaluation of our pruning approach. Finally, Section \ref{sec:future} concludes with recommendations for future research and development.

%% file: content/003_background.tex
\section{Background}
\label{sec:background}

Mamba~\cite{gu2024mambalineartimesequencemodeling} is a sequence model that replaces the quadratic-cost attention in Transformers~\cite{vaswani2023attentionneed} with selective state space models (SSMs). 
This gives linear-time complexity in sequence length while retaining strong modeling accuracy.

\begin{figure}[ht]
    \centering
    \includegraphics[width=0.65\textwidth]{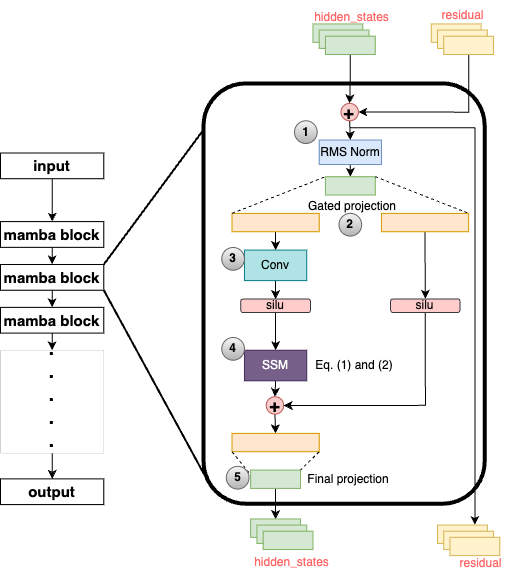}
    \caption{Mamba block. Input is normalized, gated, convolved, and passed through a selective SSM, then projected back to the hidden size with residual connections.}
    \label{fig:mamba_block}
\end{figure}

\subsection{State Space Models}
An SSM maintains a hidden state $h(t)$ that summarizes past inputs. 
It updates $h(t)$ linearly from the previous state and the current input, and produces an output $y(t)$:
\begin{align}
    h'(t) &= A h(t) + B x(t), \label{eq:state_update1}\\
    y(t) &= C h(t). \label{eq:state_update2}
\end{align}
Here, $A$ is the transition matrix, $B$ and $C$ are input/output projections, $N$ is the number of state channels, and $D$ is the hidden size.

\subsection{Discretization and Gating}
To make this computable, Mamba discretizes the equations using a learnable timescale $\Delta_t$:  
\begin{align}
    A_t &= \exp(\Delta_t A), \\
    B_t &= (\Delta_t A)^{-1}\!\left(\exp(\Delta_t A) - I\right)\Delta_t B.
\end{align}
The parameter $\Delta_t$ acts as a gate: large values down-weight past states, while small values preserve them~\cite{gaternn}.

\subsection{Mamba-2: State Space Duality}
Mamba-2 extends this with \emph{state space duality} (SSD)~\cite{jambateam2024jamba15hybridtransformermambamodels}, which provides two complementary views:
\begin{align}
    h_t &= A_t h_{t-1} + B_t x_t, \qquad 
    y_t = C_t^\top h_t, \\
    y &= (L \circ QK^\top)V.
\end{align}
The first (recurrent) form is efficient ($O(N)$), while the second (quadratic) form shows attention-like behavior. Together, SSD combines efficiency with expressiveness.

\subsection{Block Design}
Each Mamba block (Fig.~\ref{fig:mamba_block}) applies normalization and gating, then a convolution and the SSM, before projecting back to the hidden dimension with residuals. 
Variants like the multi-input SSM (MIS) and extra normalization layers~\cite{dao2024transformersssmsgeneralizedmodels,normform} improve efficiency and stability.

As the model is channelized ($N$ states) and each state is modulated by $\Delta_t$, we can measure the activity of individual channels and prune the inactive ones,  motivating our method in Section~\ref{sec:pruning}.

%% file: content/03_related_work.tex
\section{Related Work}
\label{sec:related_work}

\paragraph{Structured State Space Models}
Linear State-Space Layers (LSSLs) unify RNNs, CNNs, and continuous-time models for scalable sequence learning and strong long-range dependence modeling \cite{combiningrrnn}. More broadly, Structured State Space Models (SSMs) provide an effective alternative to Transformers via structured dynamics \cite{ssm2021}. Building on time-variant selectivity, Mamba introduces input conditioned state updates to retain or discard information efficiently, achieving state-of-the-art results across modalities with near-linear scaling \cite{gu2024mambalineartimesequencemodeling,waleffe2024empiricalstudymambabasedlanguage}. Surveys summarize Mamba’s advances and open problems \cite{qu2024surveymamba}. Mamba-2 further connects SSMs and attention through State Space Duality (SSD) and structured semiseparable matrices, yielding 2–8× speedups while remaining competitive for language modeling \cite{dao2024transformersssmsgeneralizedmodels}.

\paragraph{Network Pruning}
Pruning removes less important parameters to cut compute, memory, and energy \cite{pruning}. Complementary tooling supports profiling/monitoring of deep learning workloads and GPU performance analysis, including top-down methodologies and HPCToolkit for scalable bottleneck diagnosis \cite{Profilingandmanitoring,topdownperformance,hpctoolkit}. Unstructured pruning yields sparse weights that often require specialized hardware, whereas structured pruning removes channels/filters/layers for hardware-friendly speedups \cite{unstructuredvsstrcured}. Comprehensive surveys cover taxonomies, timing (pre-/post-training), and combinations with other compression (e.g., quantization) \cite{pruningsurvay}. Scheduling matters: policy-based early structural pruning can reduce cost while preserving accuracy \cite{whytoprune}; forward-pass-only structured pruning enables practical LLM compression without backpropagation \cite{dery2024everybodyprunenowstructured}. Structured pruning can also improve generalization \cite{xia2022structuredpruninglearnscompact}. For LLMs, formulating pruning as a Multiple Removal Problem improves post-training efficiency \cite{zhao2024pruningfoundationmodelshigh}. In SSMs, fine-grained token reduction exploiting importance and similarity boosts efficiency for models like Mamba-2 \cite{zhan2024rethinkingtokenreductionstate}. Mamba-specific compression achieves up to 1.4× inference speedup while maintaining accuracy \cite{muñoz2025mambashedderposttransformercompressionefficient}. Analyses show Mamba exhibits emergent attention-like behavior that supports long-range dependencies \cite{ali2024hiddenattentionmambamodels}; hardware co-design such as MARCA delivers reconfigurable, energy-efficient acceleration \cite{li2024marcamambaacceleratorreconfigurable}. Extensions include MambaTree for tree-structured long-range modeling \cite{xiaomambatree} and MambaSpike, which integrates spiking front-ends for low-power temporal processing \cite{qin2024mambaspikeenhancingmambaarchitecture}.

%% file: content/04_method.tex

\section{Component-Level Performance Analysis}
\label{sec:profiling}

This section evaluates Mamba components (Mamba-1 and Mamba-2) on memory cost, Time-to-First-Token (TTFT, Prefill), Time-per-Output-Token (TPOT, Decoding), and overall performance. We measure FLOPs, execution time, memory usage, and I/O parameters. All profiling runs use NVIDIA A100 (40 GB) with the PyTorch profiler \cite{PyTorchProfiler}; traces are exported as JSON and parsed. To reduce non-systematic noise, we perform three warm-up runs and then repeat measurements multiple times, reporting means. Experiments span sequence lengths around the 64 and 2048 token regimes at a fixed batch size of 8, to probe varying compute/memory regimes.

We analyze Gated MLP, Convolution, SSM Transformation, Final Projection, and Normalization—key stages in Mamba’s pipeline \cite{gu2024mambalineartimesequencemodeling}. Gated MLP supplies nonlinear feature transforms; Convolution captures local structure; SSM Transformation replaces attention for long-sequence modeling; Final Projection maps to output space; Normalization stabilizes training. For both Mamba-1 and Mamba-2, we extract per-layer time and memory by matching kernel launches to GPU allocations; the script iterates trace events to aggregate layer metrics. Total FLOPs are estimated from theoretical counts of matrix multiplies, activations, and convolutions \cite{efficienymisnomer}, using the HuggingFace implementations of Mamba-1/2 \cite{huggingface_transformers}.

Our goal is to locate optimization opportunities—e.g., eliminating redundant computation, refactoring memory access, and improving parallel execution. From component-level cost profiles, we identify bottlenecks and candidate improvements; the following subsections report detailed results and their implications for Mamba efficiency.

\subsection{Latency Analysis}

\begin{figure}[t]
    \centering
    
    \begin{subfigure}[b]{0.9\linewidth}
        \centering
        \includegraphics[width=\linewidth]{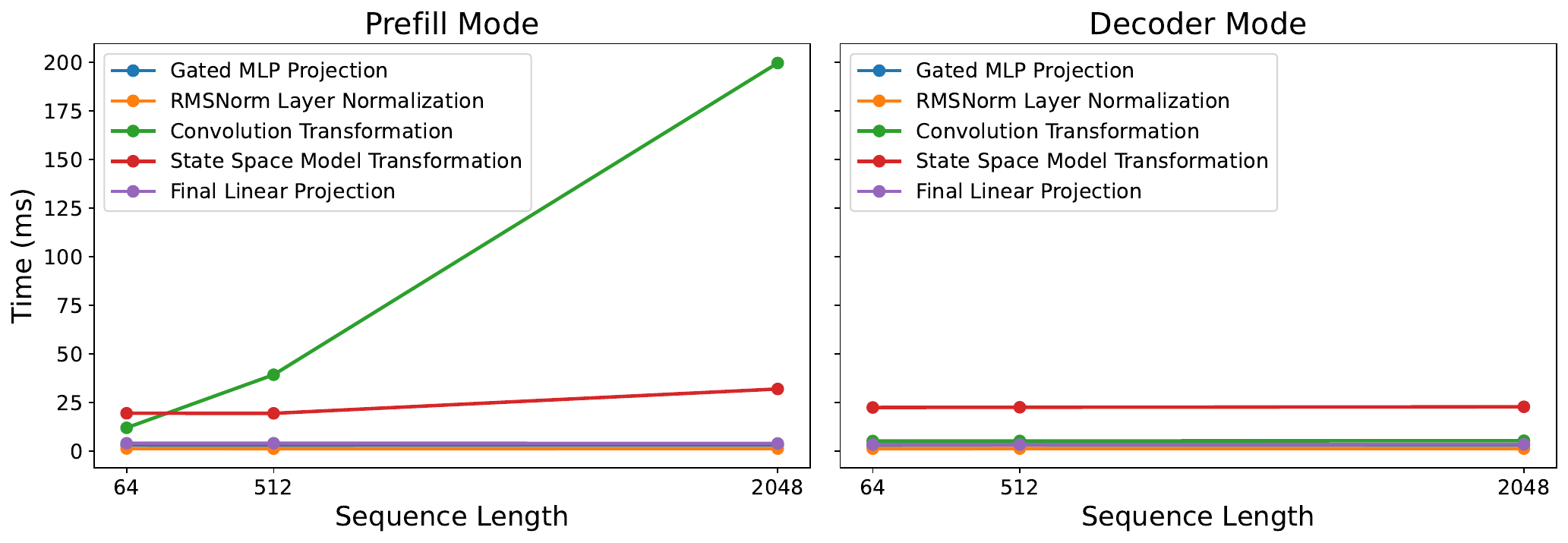}
        \caption{Latency breakdown for Mamba-1}
        \label{fig:prefil_m1}
    \end{subfigure}
    
    \vspace{0.1cm} 
    
    \begin{subfigure}[b]{0.9\linewidth}
        \centering
        \includegraphics[width=\linewidth]{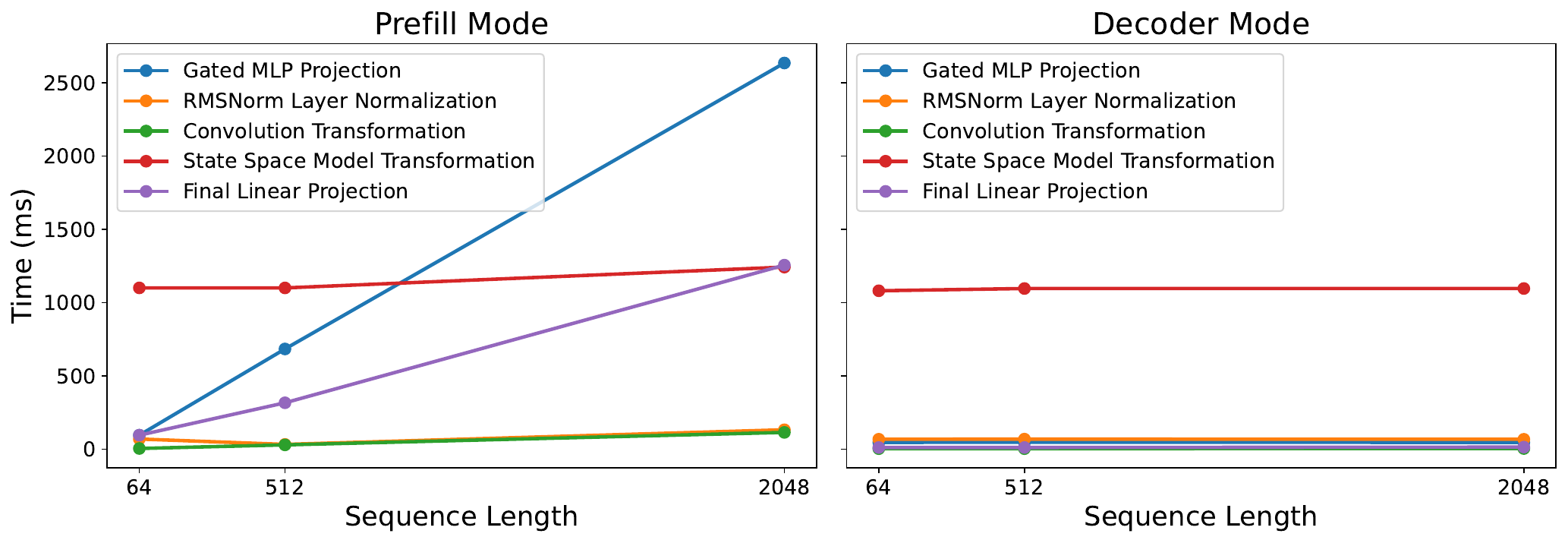}
        \caption{Latency breakdown for Mamba-2}
        \label{fig:prefil_m2}
    \end{subfigure}
    \vspace{-0.8em}
    \caption{Latency breakdown of individual components in Mamba-1 and Mamba-2 across Prefill and Decoder modes.}
    \label{fig:prefill_combined}
\end{figure}

Latency in Mamba arises in two modes: \emph{Prefill} and \emph{Decoder}. Prefill processes the entire sequence at once to initialize states, while Decoder generates tokens one by one using these states. Prefill is length-dependent and costly; Decoder is length-independent and dominates real-world inference such as autoregressive generation.

In Prefill, Mamba-1 is bottlenecked by the convolution:
\[
y_t = \sum_{i=0}^{k-1} W_i x_{t-i},
\]
which scales as $O(kL^2)$ with sequence length $L$ due to sequential dependencies (Fig.~\ref{fig:prefil_m1}).  
Mamba-2 reduces this cost using 2D scanning,
\[
y_t = \text{Conv2D}(X_{t,s}) + \text{SSM}(\text{Scan}(X_{t,s})),
\]
achieving $O(kL)$ complexity (Fig.~\ref{fig:prefil_m2}). The main prefill bottleneck then shifts to the gated MLP, where
\[
[z_x \,\Vert\, B \,\Vert\, C \,\Vert\, \Delta_t] = W_{\text{proj}}x_t,
\]
requires $O(\gamma DN L)$ operations, with state dimension $N$ and expansion factor $\gamma$.

In Decoder mode, costs change fundamentally. Convolution and projection overheads vanish because only one new token is processed at a time. Each step requires a single SSM recurrence,
\[
h_t = A_t h_{t-1} + B_t x_t,
\]
whose $O(N)$ cost is independent of $L$. As a result, the SSM is always the dominant component in Decoder latency. Other layers (normalization, output projection) add constant, negligible overhead.  

\textbf{In summary}, Prefill bottlenecks differ by architecture (convolution in Mamba-1, gated MLP in Mamba-2), but in Decoder mode—the critical path for practical inference—the SSM consistently dominates latency. This makes SSM optimization the most effective route to real-world speedups.

\subsection{Memory Analysis}
Memory usage in Mamba is dominated by the State Space Model (SSM) component. At a sequence length of 2048, Mamba-2’s SSM consumes 33.5\% more memory than Mamba-1 (115.68 GB vs. 86.64 GB) due to block-wise state materialization, which improves cache efficiency but increases memory demand. Localized state transitions in Mamba-2 reduce redundant updates, partly offsetting this cost.  
The Gated MLP in Mamba-2 achieves 11.6\% lower memory usage (10.32 GB vs. 11.52 GB) via parallel parameter generation, optimizing layout and reducing overhead for large sequences.  
Smaller layers such as RMSNorm scale predictably with \( O(BLD) \), e.g., from 0.5 GB at length 64 to 15.75 GB at 2048 in Mamba-2, mainly for per-channel statistics. While minor in overall usage, they help explain scaling trends in larger components.

\input{table/t01_combine}

\subsection{Computational Complexity Analysis}
The SSM is the most computationally intensive component, dominating FLOPs across both architectures. At a sequence length of 2048, Mamba-2 reduces FLOPs by 28.6\% compared to Mamba-1 (1566.55G vs. 2193.73G) through block-wise processing, which minimizes redundant matrix multiplications and improves scalability without sacrificing throughput. The Gated MLP also benefits from parallel parameter projection, resulting in a 20.3\% reduction in FLOPs at 2048 tokens and improved GPU utilization.  
Other layers, including RMSNorm and Convolutional Transformations, contribute only 5–10\% of total FLOPs and remain consistent across models. These results underscore how targeted design choices in the SSM and Gated MLP significantly enhance computational efficiency.

\subsection{Data Flow and Hardware Utilization}
I/O bandwidth analysis reveals how computational components interact with memory hierarchies and GPU resources. The State Space Model (SSM) has the highest I/O demand due to recurrent state updates for long-range dependencies. Unlike standard matrix multiplications, its sequential data dependencies cause complex memory access patterns. Block-wise state transfers group states into contiguous regions, improving cache efficiency and reducing global memory fetches. This enables peak bandwidth of 9.87 GB/s at sequence length 2048—a 26.3\% gain over a sequential layout—showing that localized state materialization better leverages GPU memory.

The Gated MLP, responsible for token embedding transformations, exhibits dense parameter reads and processes tokens independently, enabling parallelization. Its I/O bandwidth increases by over 51\% for long sequences, underscoring the role of parallel parameter loading in minimizing stalls.
\begin{equation}
P_{\text{parallel}} =
\begin{bmatrix}
x_t \parallel \Delta_t \parallel B_t \parallel C_t
\end{bmatrix}
\in \mathbb{R}^{B \times L \times (D + 3N)}
\end{equation}
Here, \(x_t\) are input embeddings, \(\Delta_t\) the learned time-step scalars, and \(B_t, C_t\) the selective gating parameters; \(B\), \(L\), \(D\), and \(N\) denote batch size, sequence length, hidden dimension, and state channels. Concatenating these allows simultaneous retrieval, exploiting GPU memory coalescing and reducing latency. RMSNorm, though minor in footprint, scales predictably with \(O(BLD)\) and provides a baseline for I/O efficiency. Despite having few parameters, the SSM still dominates FLOPs, memory, and I/O usage, as confirmed by manual analysis from official code. Results may vary with GPU configurations.

%% file: table/t01_combine.tex
\setlength{\tabcolsep}{6pt} 
\renewcommand{\arraystretch}{1.0} 

\begin{table*}[]
\centering
\small
\caption{Component-wise Resource Utilization in Mamba-1 and Mamba-2 Architectures}
\label{tab:resource_usage_combined}
\resizebox{\textwidth}{!}{%
\begin{tabular}{p{1cm}|l|cccccccc}
\toprule
\multirow{2}{*}{\textbf{Seq}} & \multirow{2}{*}{\textbf{Component}} & \multicolumn{2}{c}{\textbf{FLOPs (G)}} & \multicolumn{2}{c}{\textbf{Memory (GB)}} & \multicolumn{2}{c}{\textbf{I/O (GB/s)}} & \multicolumn{2}{c}{\textbf{Latency (ms)}} \\
\cmidrule(lr){3-4} \cmidrule(lr){5-6} \cmidrule(lr){7-8} \cmidrule(lr){9-10}
 & & M-1 & M-2 & M-1 & M-2 & M-1 & M-2 & M-1 & M-2 \\
\midrule
\multirow{5}{*}{64} & RMSNorm & 0.06 & 0.06 & 0.75 & 0.5 & 0.0016 & 0.08 & 3.20 & 2.90 \\
 & Gated MLP & 2.42 & 29.03 & 0.24 & 0.48 & 0.0039 & 0.45 & 4.08 & 3.99 \\
 & Conv. Transform & 0.60 & 0.26 & 0.48 & 0.24 & 0.0032 & 0.19 & 9.40 & 0.21 \\
 & State Space & 68.63 & 48.95 & 6.24 & 7.44 & 0.042 & 0.31 & 16.76 & 45.88 \\
 & Final Linear & 28.99 & 14.50 & 0.96 & 0.48 & 0.0024 & 0.23 & 3.13 & 2.18 \\
\midrule
\multirow{5}{*}{512} & RMSNorm & 0.45 & 0.45 & 6.25 & 4.0 & 0.01 & 0.6 & 2.02 & 1.37 \\
 & Gated MLP & 19.37 & 232.23 & 2.64 & 0.2 & 0.03 & 1.87 & 26.16 & 28.51 \\
 & Conv. Transform & 4.83 & 2.11 & 1.92 & 2.16 & 0.03 & 1.51 & 9.59 & 1.23 \\
 & State Space & 548.49 & 391.64 & 22.08 & 29.76 & 0.34 & 2.47 & 23.73 & 35.87 \\
 & Final Linear & 231.93 & 115.96 & 7.44 & 2.88 & 0.02 & 1.02 & 15.12 & 13.20 \\
\midrule
\multirow{5}{*}{2048} & RMSNorm & 1.81 & 1.81 & 25.18 & 15.75 & 0.05 & 2.42 & 7.65 & 5.53 \\
 & Gated MLP & 77.45 & 928.92 & 10.32 & 11.52 & 0.13 & 6.73 & 98.58 & 109.83 \\
 & Conv. Transform & 19.33 & 8.46 & 7.44 & 8.4 & 0.10 & 6.04 & 12.81 & 4.78 \\
 & State Space & 2193.73 & 1566.55 & 86.64 & 115.68 & 1.35 & 9.87 & 74.82 & 51.79 \\
 & Final Linear & 927.71 & 463.86 & 23.04 & 5.04 & 0.08 & 3.74 & 59.99 & 52.35 \\
\bottomrule
\end{tabular}
}
\end{table*}

%% file: content/05_SSM.tex
\section{Pruning Mamba's State Representation}
\label{sec:pruning}

Profiling in Section~\ref{sec:profiling} identified the State Space Model (SSM) component as the primary computational bottleneck in both Mamba-1 and Mamba-2, accounting for 60–70\% of total runtime and memory usage. Mamba-1 implements SSM via the Selective Structured State Space (S6) formulation, while Mamba-2 enhances this with the State Space Duality (SSD) framework~\cite{jambateam2024jamba15hybridtransformermambamodels}. Both variants employ a selective update mechanism for sequence processing, as described in Eqs.~\eqref{eq:state_update1} and \eqref{eq:state_update2}.

\noindent The gating factor $\Delta_t$ is computed as:
\begin{equation}
\Delta_t = \text{softplus}(W_\Delta x_t),
\end{equation}
and controls the balance between retaining the previous state $h_{t-1}$ and incorporating new input $x_t$~\cite{gu2024mambalineartimesequencemodeling}. 
 In practice, Mamba parameterizes $A$ \emph{diagonally} (state-wise) for efficiency; thus $A_t=\exp(\Delta_t A)$ is computed as per-state exponentials $a_{t,s}=\exp(\Delta_t a_s)$ rather than a dense matrix exponential~\cite{gu2024mambalineartimesequencemodeling}. 
Given that $A$ is initialized with negative entries, large $\Delta_t$ values drive the corresponding exponentials toward zero, suppressing the recurrent term $A_t h_{t-1}$, allowing $B_t = \Delta_t B$ to dominate and yielding $h_t \approx B_t x_t$ (current input dominance). Conversely, when $\Delta_t$ is small, $A_t \approx I$ and $B_t \approx 0$, so $h_t \approx h_{t-1}$, preserving past state information with minimal new input influence.

Here, $A_t \in \mathbb{R}^{N \times N}$ is the state transition, $B_t \in \mathbb{R}^{N \times 1}$ is the input projection, $C \in \mathbb{R}^{1 \times N}$ is the output projection, and $N = d_{\text{state}}$ denotes the number of \emph{state channels}. The hidden state vector $h_t \in \mathbb{R}^N$ consists of $N$ scalar components, each representing one state $s$. A ``state'' in this context corresponds both to the $s$-th component of $h_t$ and to the associated row and column in $A_t$, as well as the $s$-th entries in $B_t$ and $C$.

For each layer $l$, we maintain a \emph{state activity matrix}:
\[
S_l \in \mathbb{R}^{d_{\text{state}} \times n_{\text{samples}}},
\]
where $d_{\text{state}} = N$ and $n_{\text{samples}}$ equals the total number of time steps evaluated (sequence length $L$ times the number of sequences processed during profiling). The average activity of state $s$ in layer $l$ is:
\begin{equation}
\text{activity}(l,s) = \mathbb{E}_{x \sim \mathcal{D}}\left[\Delta_t^{(l,s)}\right],
\end{equation}
where $\Delta_t^{(l,s)}$ is the $s$-th entry of the $\Delta_t$ vector in layer $l$. In Mamba-2’s multi-head configuration, states are grouped into clusters of size $d_{\text{state}} / n_{\text{heads}}$ per head, enabling head-specific importance analysis (Section~\ref{sec:background}).

\begin{figure}[ht]
    \centering
    \includegraphics[width=0.7\textwidth]{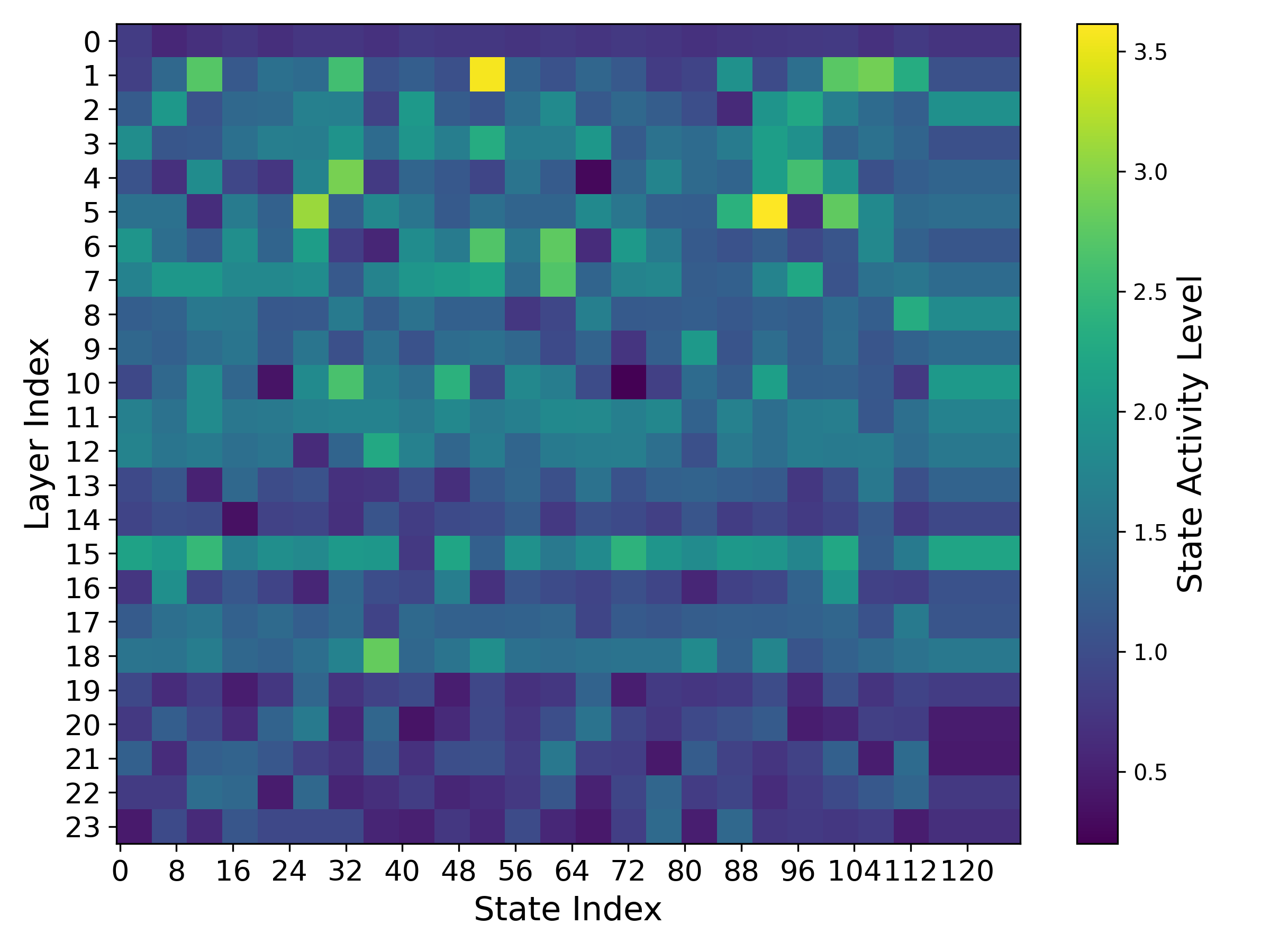}
    \caption{State Importance Heatmap for Mamba2-130M: brighter colors indicate higher average $\Delta_t$ activity.}
    \label{fig:layer_heatmap}
\end{figure}

Figure~\ref{fig:layer_heatmap} shows that certain states maintain low activity across most inputs, especially in early and late layers, while others are consistently high-activity. This suggests redundancy that can be exploited through pruning.

\paragraph{Pruning approach.}

From Eq.~\eqref{eq:state_update1}, removing a state $s$ entails zeroing the $s$-th row and column of $A_t$ and the corresponding entries in $B_t$, $C$, and $h_t$. This effectively eliminates the recurrent and input/output contributions of that state without requiring the entire $A_t$ to vanish. As states are parameterized independently by channel, pruning operates at the \emph{channel level} and removes full $(A,B,C)$ contributions for state $s$, avoiding the ambiguity of element-wise masking.

\input{table/algorithm_v2}

Algorithm~\ref{alg:ssm_pruning} selects the $k = \lfloor N(1-r) \rfloor$ most active channels for a pruning ratio $r$, using the activity scores from Eq.~(10). 
After pruning, the state dimension is reduced from $N$ to $N'$, and the input projection $W_{\text{proj}}$ is updated so that it no longer produces parameters for the removed channels. 
Because later layers are implemented to expect the original size $N$, we add a \emph{bridging layer} $W_{\text{bridge}} \in \mathbb{R}^{N \times N'}$. 
This layer takes the reduced hidden state $h_{t,\text{keep}} \in \mathbb{R}^{N'}$ and maps it back to $h_{t,\text{bridge}} \in \mathbb{R}^N$. 
In this way, normalization and output projections can operate without any change to their dimensions.

We evaluate three configurations: (1) a dense baseline; (2) a sparse variant with pruned states zeroed but retained in memory; and (3) an optimized variant with pruned states physically removed, reducing FLOPs and memory usage. All evaluations use a 10-iteration warm-up followed by 100 measured iterations with explicit CUDA synchronization to minimize measurement variance.

%% file: table/algorithm_v2.tex

\begin{algorithm}
\caption{Activity-Based SSM State Selection (per layer $l$)}
\label{alg:ssm_pruning}
\begin{algorithmic}[1]
\State \textbf{Input:} input $X \in \mathbb{R}^{B \times L \times D}$, 
SSM parameters $(A,B,C,\Delta)$, state dimension $N$, pruning ratio $r$
\State \textbf{Output:} pruned hidden state $h_{t,\text{bridge}} \in \mathbb{R}^{N}$

\State \textbf{Step 1: Score states}
\State Initialize activity scores $s \in \mathbb{R}^N$
\For{$s = 1 \ldots N$}
    \State Compute gating values $\Delta_t^{(s)}$ for $X$ \hfill (Eq.~9)
    \State $s[s] \gets \mathbb{E}[\Delta_t^{(s)}]$ \hfill (Eq.~10)
\EndFor

\State \textbf{Step 2: Select active states}
\State $k \gets \lfloor N(1-r) \rfloor$
\State $I_{\text{keep}} \gets \text{Top-}k(s)$
\State \Return $I_{\text{keep}}$
\end{algorithmic}
\end{algorithm}

%% file: content/06_eval.tex
\section{Evaluation and Results}
\label{sec:results}

\subsection{Experimental Setup}

We evaluate pruning on the Mamba2-130M model, chosen for its improved SSM state management, computational efficiency, and balanced capacity–cost trade-off. This scale enables detailed state dynamics analysis, scales to other model sizes, and supports extensive pruning experiments within practical resource limits. Zero-shot performance is measured on four diverse benchmarks—PIQA~\cite{bisk2019piqareasoningphysicalcommonsense}, Arc Easy~\cite{clark2018thinksolvedquestionanswering}, Hellaswag~\cite{zellers2019hellaswagmachinereallyfinish}, and OpenBookQA~\cite{sun2019improvingmachinereadingcomprehension}—without task-specific fine-tuning, assessing reasoning and comprehension under varying pruning ratios.

\subsection{Impact of Pruning on Model Accuracy}
The baseline (unpruned) model demonstrates varying levels of competence in performing diverse tasks. Among these, PIQA has the highest baseline accuracy of 65\%, followed by Arc Easy with 57\%, Hellaswag with 37\%, and OpenBookQA with 22\%. The variation in baseline performance highlights the intrinsic challenge and reasoning requirements of each task. Under increasingly stricter pruning, each task exhibits a particular trend of degradation. PIQA remains the most robust, achieving high performance despite aggressive pruning, dropping by only five percentage points (to 60\%) at a pruning ratio of 0.9. Arc Easy is the most susceptible to pruning, with its accuracy dropping significantly from 57\% to 25\% at a 0.9 pruning ratio. Hellaswag shows a consistent but modest decline from 37\% to 28\%, while OpenBookQA is remarkably stable, remaining nearly constant at around 20--22\% across all pruning ratios.

\begin{figure}[ht]
    \centering

    \includegraphics[width=1\textwidth]{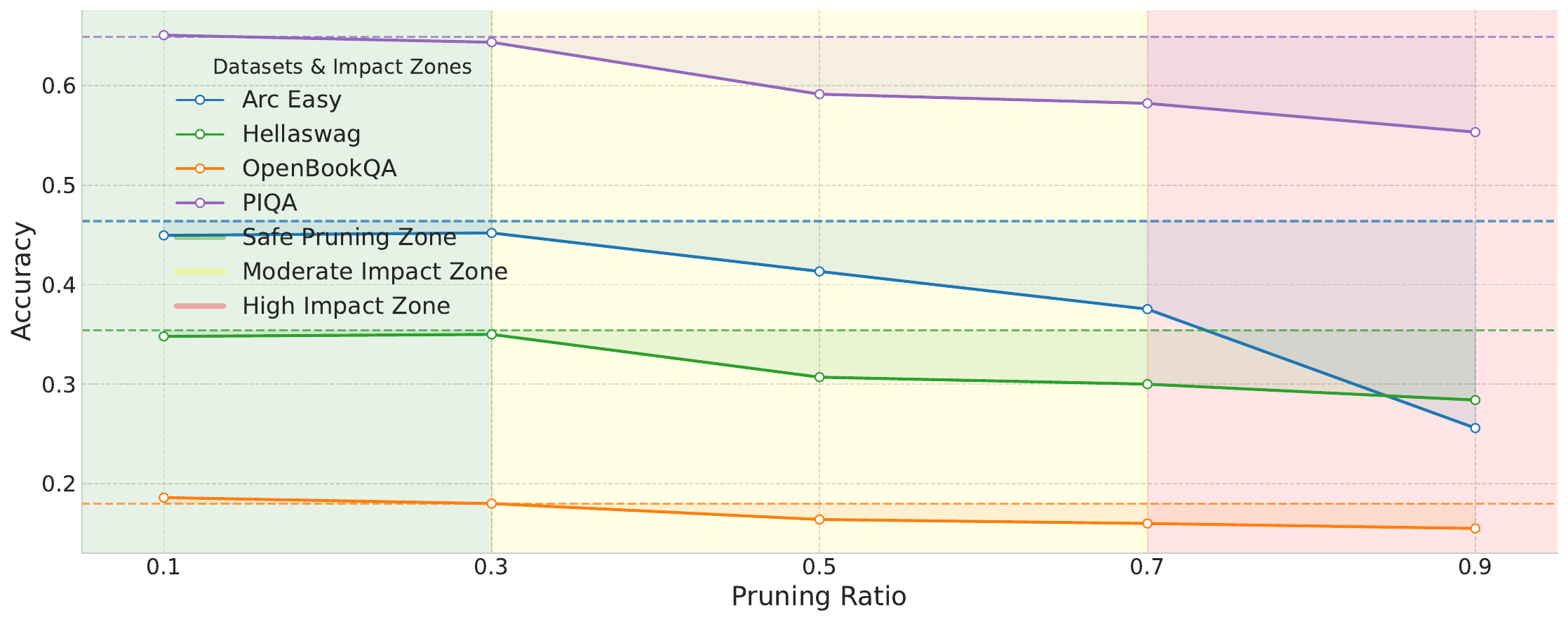} 
    \caption{Effect of state pruning on model accuracy across various datasets. 
    Dotted lines indicate baseline accuracies, while solid lines represent the performance of the pruned model. The background is divided into three pruning zones based on average performance degradation.}
    \label{fig:acc_dataset}
\end{figure}

Based on the observed performance patterns, we recognized three distinct operational regions that characterize the effect of pruning:
\begin{itemize}
    \item \textbf{Safe Pruning Region} ($\leq 0.3$ pruning ratio): This is the range within which the model's performance is almost as good as the baseline for every task, with a mean accuracy loss of merely 0.7\%. This region implies that roughly 30\% of the model's states can be pruned without much performance loss. Therefore, it reflects a high computational redundancy in the original model.
    \item \textbf{Moderate Impact Zone} ($0.3$--$0.7$ pruning ratio): This zone marks the onset of quantifiable drops in performance, with a mean 12.2\% reduction in accuracy for all tasks. The different breakdown rates observed across this zone, ranging from no visible effect on PIQA to precipitous declines for Arc Easy, illustrate task-specific dependencies for model capacity and computational heuristics.
    \item \textbf{High Impact Zone} (pruning ratio $> 0.7$): This zone demonstrates pruning conditions of high impact that result in extensive performance degradation for most tasks, as expressed by an average accuracy loss of 22.3\%. Still, the stark contrasts in degradation trends, from PIQA's high resilience to Arc Easy's sharp decline, provide insight into the spread of task-specific computational demands across different model states.
\end{itemize}
This zoning analysis is critical for deploying pruned models in high-performance computing environments. The Safe Zone defines clear boundaries for optimization without performance degradation, and the Moderate and High Impact Zones provide flexible trade-offs between computational efficiency and task-specific performance needs. These observations are particularly useful in resource-constrained deployments that require the optimal balance between computation and accuracy.

\subsection{Resource Efficiency Improvements}

\begin{figure*}[t]
    \centering
    \begin{subfigure}[b]{0.49\textwidth}
        \centering
        \includegraphics[width=\linewidth]{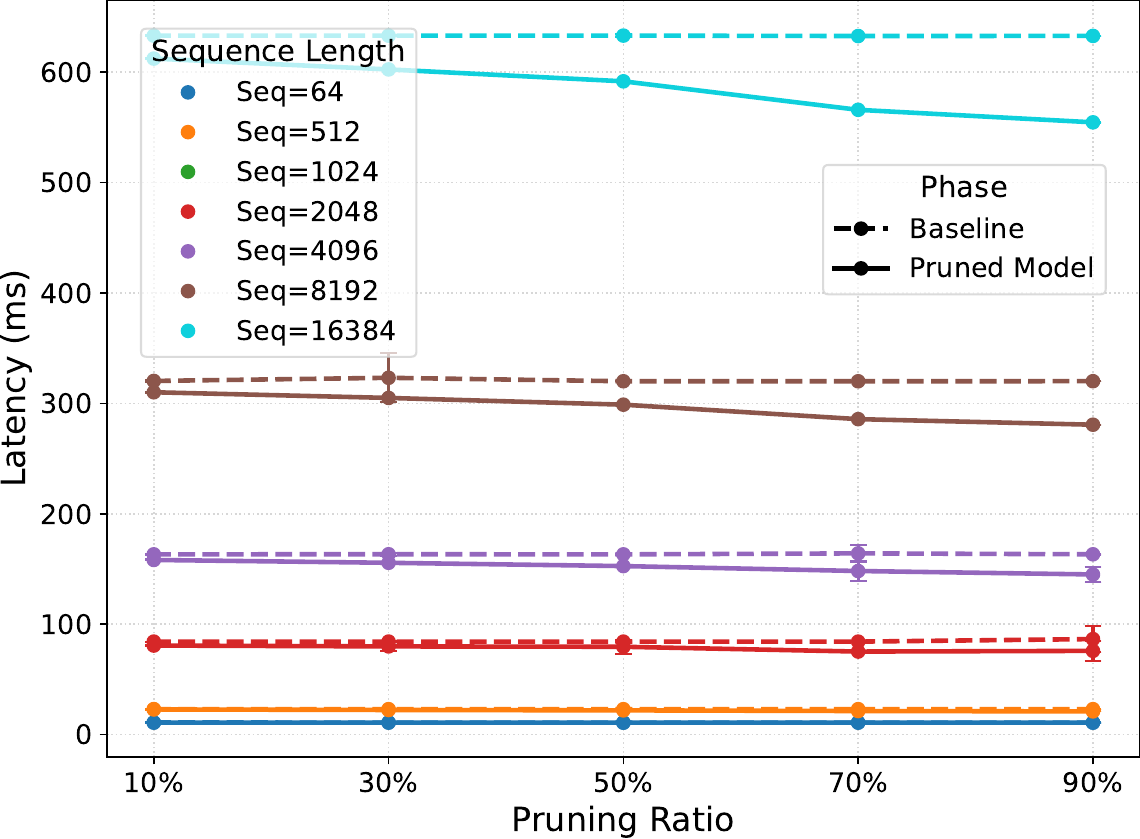}
        \caption{Latency versus pruning ratio for various sequence lengths.}
        \label{fig:sim_latency}
    \end{subfigure}
    \hfill
    \begin{subfigure}[b]{0.49\textwidth}
        \centering
        \includegraphics[width=\linewidth]{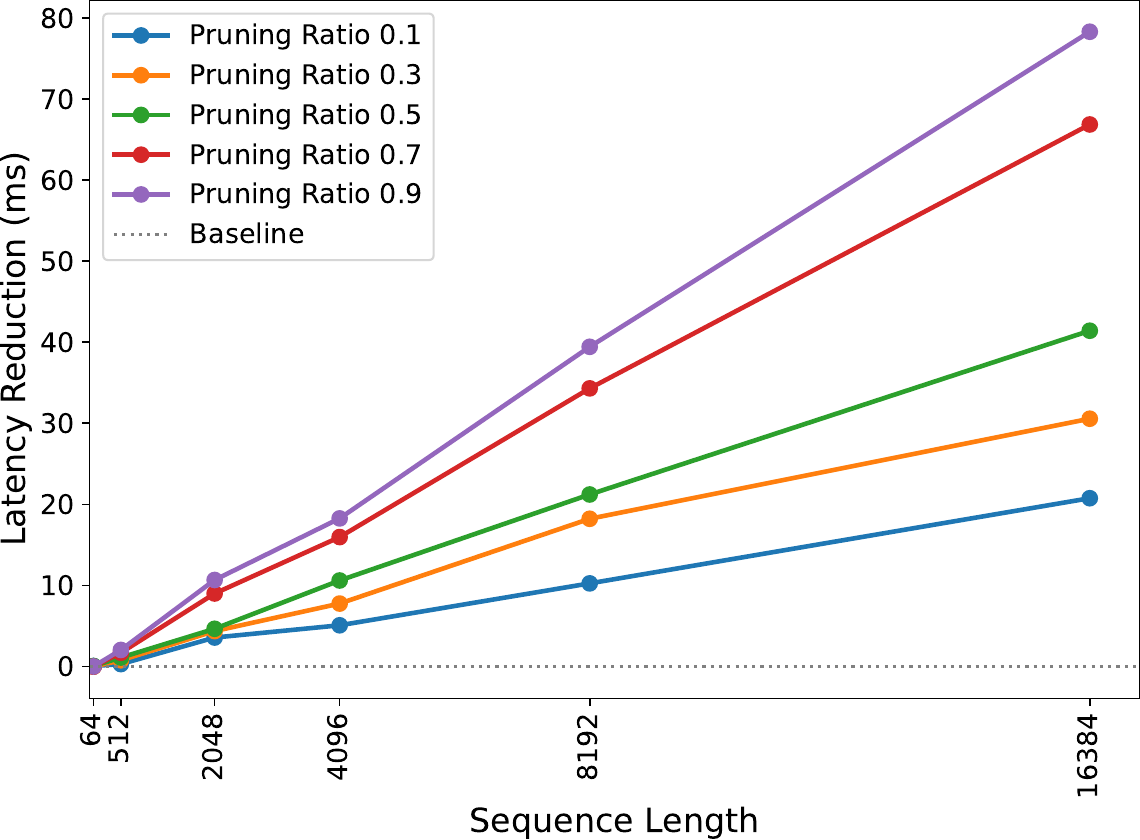}
        \caption{Latency reduction across sequence lengths compared to baseline.}
        \label{fig:latency_diff}
    \end{subfigure}
    \caption{(a) Latency versus pruning ratio and (b) Latency reduction across sequence lengths.}
    \label{fig:combined}
\end{figure*}

To evaluate the efficiency gains from our pruning method, we performed latency measurements across sequence lengths of 64 to 16,384 tokens and pruning ratios of 0.1 to 0.9. Our findings reveal a complex relationship between sequence length and pruning effectiveness (in Figure \ref{fig:sim_latency}). For short sequences ($\leq 512$ tokens), pruning minimally affects latency, with base times around 15 ms for 64 tokens and 45 ms for 512 tokens. This suggests that state management overhead is negligible for short sequences. The impact of pruning becomes more evident for medium-length sequences (2,048 to 8,192 tokens). For 4,096-token sequences, latency decreases from 300 ms to 280 ms as the pruning ratio increases to 0.9. Similarly, for 8,192-token sequences, latency improves from 400 ms to 370 ms. The most significant gains occur with longer sequences (16,384 tokens), where pruning reduces latency from 630 ms to 560 ms at a pruning ratio of 0.9, resulting in approximately an 11\% improvement. This suggests that pruning is more effective with longer sequences, where computational savings are more pronounced. Overall, the effectiveness of state pruning increases with sequence length, indicating that pruning strategies should be dynamically adjusted to optimize resource utilization and processing efficiency in distributed computing scenarios.

\input{table/speedup_mem_table}

To showcase the significant benefits of our pruning method in high-performance computing, we evaluated computational speedup and memory savings across various sequence lengths and pruning ratios in Table \ref{tab:speedup_memory}. For very short sequences (64 tokens), our method achieves memory reduction of up to 7.82\%  at a 0.9 pruning ratio, demonstrating efficient memory optimization even in scenarios where computational overheads dominate. As sequence length increases, the advantages of pruning become even more pronounced. For medium lengths (512--2048 tokens), we observe consistent speedup ratios ranging from 1.10$\times$ to 1.14$\times$, accompanied by substantial memory savings of 10.30\% to 10.97\%. These results highlight the growing importance of state management and the effectiveness of our method in optimizing longer sequences. Our method delivers sustained improvements for longer sequences (4096 tokens to 16,384 tokens), achieving speedup ratios of up to 1.14$\times$ and memory savings of 11.50\% . This plateauing at a high level of optimization reflects the robustness of our approach, which maximizes computational efficiency without compromising model performance. Notably, the relationship between pruning ratio and efficiency gains is carefully balanced. While aggressive pruning (0.7--0.9) demonstrates diminishing returns, it further underscores the versatility of our method in maintaining a balance between efficiency and model accuracy. Overall, Table \ref{tab:speedup_memory} highlights these compelling trends in resource utilization, demonstrating that pruning becomes increasingly impactful as sequence lengths grow, making it a valuable tool for high-performance distributed computing scenarios. 



\subsection{Accuracy--Latency Trade-off}

\begin{figure}[ht]
    \centering
    \includegraphics[width=1\textwidth]{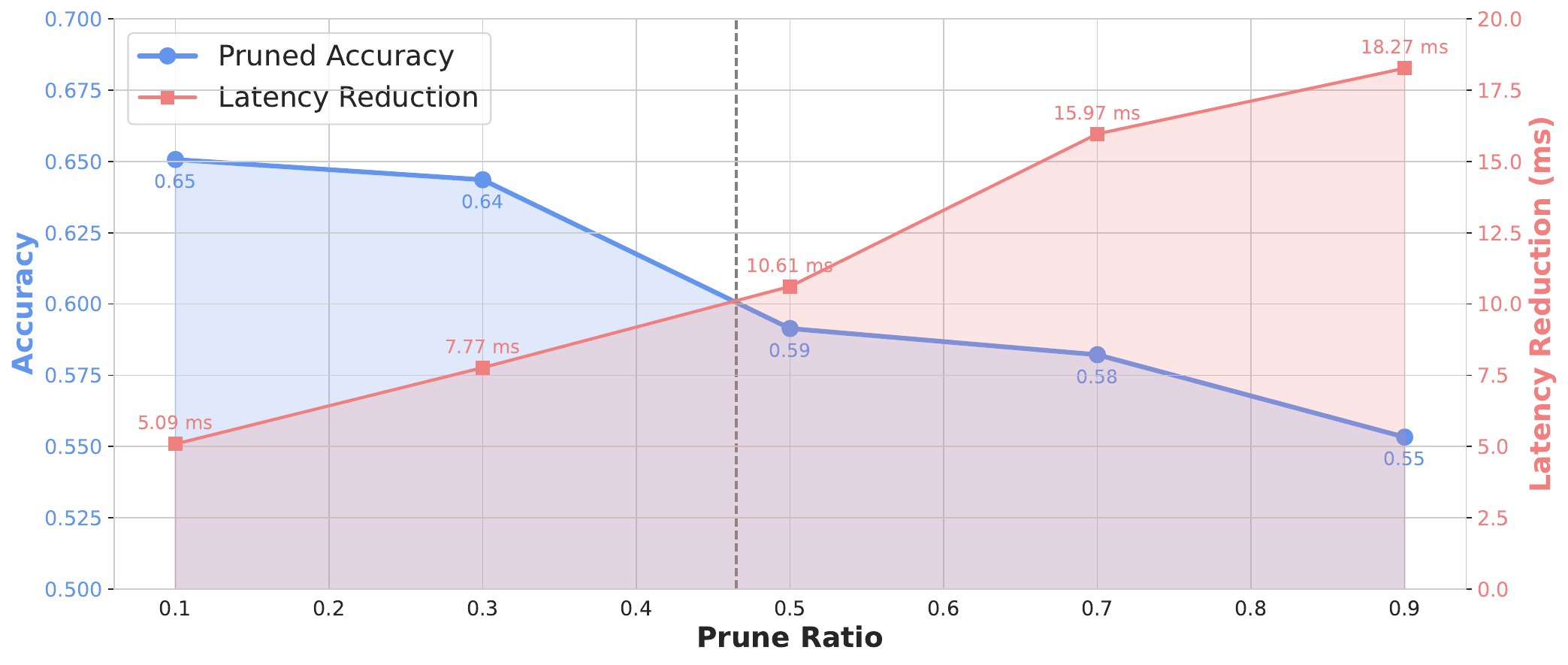}
    \caption{Accuracy-latency trade-off across pruning ratios, highlighting optimal balance at moderate pruning levels for efficient deployment.
}
    \label{fig:tradeoff}
    \vspace{-1.5em}
\end{figure}

Figure~\ref{fig:tradeoff} illustrates the accuracy--latency behavior across pruning ratios. For pruning up to $0.4$--$0.5$, latency decreases steadily (5.09\,ms to 10.61\,ms) while accuracy remains relatively stable (0.65 to 0.60), indicating removal of mostly redundant states. Beyond this point, accuracy drops more sharply (0.59 to 0.55) despite further latency gains (reaching 18.27\,ms), showing that pruning starts affecting states essential for selective scanning and long-range mixing. This identifies $\sim$0.5 as a practical threshold: lower ratios suit accuracy-critical tasks, whereas higher pruning favors throughput-oriented or resource-limited deployments.

To quantify this shift, we use the marginal accuracy drop,
\[
\Delta_{\text{acc}} = \frac{\text{Acc}(r+\delta)-\text{Acc}(r)}{\delta},
\]
where $r$ is the pruning ratio and $\delta$ is a small increment (here, $0.1$). Empirically, $\lvert \Delta_{\text{acc}} \rvert < 0.01$ for $r\leq0.5$ but exceeds $0.03$ beyond it, confirming nonlinear degradation past the threshold. 
\vspace{-1.5em}

%% file: table/speedup_mem_table.tex
\begin{table*}[t]
    \centering
    \small 
    \renewcommand{\arraystretch}{0.85} 
    \caption{Impact of Pruning on Computational Efficiency: The table presents the speedup factor (×) and memory reduction (\%) for different sequence lengths and pruning ratios. The speedup factor quantifies the relative decrease in latency compared to the baseline model, while the memory reduction percentage indicates the relative decrease in memory usage.}
    \label{tab:speedup_memory}
    \resizebox{\linewidth}{!}{%

    \begin{tabular}{p{1.2cm}|l|ccccc} 
        \toprule
        \multirow{2}{*}{\centering \textbf{Seqlen}} & \multirow{2}{*}{\textbf{Metric}} & \multicolumn{5}{c}{\textbf{Pruning Ratio}} \\  
        \cmidrule{3-7} 
         &  & \textbf{0.1} & \textbf{0.3} & \textbf{0.5} & \textbf{0.7} & \textbf{0.9} \\  
        \midrule
        \multirow{2}{*}{\centering 64}    & Speedup (×)    & 1.01× & 1.00× & 1.01× & 1.00× & 1.00× \\
                                          & Mem. Red. (\%) & 0.82\% & 2.58\% & 4.35\% & 6.05\% & 7.82\% \\
        \midrule
        \multirow{2}{*}{\centering 512}   & Speedup (×)    & 1.01× & 1.03× & 1.05× & 1.08× & 1.10× \\
                                          & Mem. Red. (\%) & 0.84\% & 2.07\% & 4.79\% & 7.99\% & 10.30\% \\
        \midrule
        \multirow{2}{*}{\centering 2048}  & Speedup (×)    & 1.04× & 1.05× & 1.06× & 1.12× & 1.14× \\
                                          & Mem. Red. (\%) & 0.88\% & 3.45\% & 6.08\% & 8.31\% & 10.97\% \\
        \midrule
        \multirow{2}{*}{\centering 4096}  & Speedup (×)    & 1.03× & 1.05× & 1.07× & 1.11× & 1.13× \\
                                          & Mem. Red. (\%) & 1.32\% & 3.66\% & 6.21\% & 8.87\% & 11.20\% \\
        \midrule
        \multirow{2}{*}{\centering 8192}  & Speedup (×)    & 1.03× & 1.06× & 1.07× & 1.12× & 1.14× \\
                                          & Mem. Red. (\%) & 1.18\% & 3.82\% & 6.35\% & 8.88\% & 11.50\% \\
        \midrule
        \multirow{2}{*}{\centering 16384} & Speedup (×)    & 1.03× & 1.05× & 1.07× & 1.12× & 1.14× \\
                                          & Mem. Red. (\%) & 1.40\% & 3.59\% & 6.38\% & 8.78\% & 11.47\% \\
        \bottomrule
    \end{tabular}
  }  
\end{table*}

%% file: content/07_future_work.tex
\section{Conclusion and Future work}
\label{sec:future}


This work profiled the Mamba architecture and showed that the state space module is the dominant bottleneck in both latency and memory. We introduced an activity-based pruning approach that reduces redundant states and improves efficiency while preserving functionality. Future directions include dynamic cache allocation, cross-layer cache sharing, hardware-optimized sparse-state computation, and extending the pruning evaluation to larger datasets and longer-sequence workloads to validate scalability.

%% file: content/08_ack.tex
\subsubsection*{Acknowledgments.}
We extend our gratitude to Intel Labs for supporting this project. This work utilized the Delta system at the National Center for Supercomputing Applications (NCSA) through allocation CIS240626. We also acknowledge support from the National Science Foundation under grants 2138259, 2138286, 2138307, 2137603, and 2138296.
